\documentclass{article}
\usepackage[utf8]{inputenc}
\usepackage{physics}
\usepackage{times}
\usepackage{microtype}
\usepackage{graphicx}
\usepackage{xcolor,colortbl}
\usepackage{amssymb}
\usepackage{soul}
\usepackage{subcaption}
\usepackage{booktabs} 
\usepackage{algorithm,algpseudocode}
\usepackage{multirow}
\usepackage{amsmath}
\usepackage{comment}
\usepackage{caption}
\usepackage[T1]{fontenc}
\usepackage{authblk}
\usepackage{bm}
\DeclareMathOperator*{\argmax}{arg\,max}
\DeclareMathOperator*{\argmin}{arg\,min}

\title{Parametric Noise Injection: Trainable Randomness to Improve Deep Neural Network Robustness against Adversarial Attack}

\author[1]{Adnan Siraj Rakin and Zhezhi He \thanks{Equally contributed}}
\author[1]{Deliang Fan\thanks{Corresponding Author: dfan@ucf.edu}}
\affil[1]{Department of Computer Engineering, University of Central Florida}

\date{}

\usepackage{hyperref}

\usepackage{import}
\usepackage{tikz}
\usetikzlibrary{positioning}

\usepackage[capitalise]{cleveref}

\usepackage{geometry}

\begin{document}

\maketitle

\begin{abstract}
Recent development in the field of Deep Learning have exposed the underlying vulnerability of Deep Neural Network (DNN) against adversarial examples. In image classification, an adversarial example is a carefully modified image that is visually imperceptible to the original image but can cause DNN model to misclassify it. Training the network with Gaussian noise is an effective technique to perform model regularization, thus improving model robustness against input variation. Inspired by this classical method, we explore to utilize the regularization characteristic of noise injection to improve DNN's robustness against adversarial attack. In this work, we propose Parametric-Noise-Injection (PNI) which involves trainable Gaussian noise injection at each layer on either activation or weights through solving the min-max optimization problem, embedded with adversarial training. These parameters are trained explicitly to achieve improved robustness. To the best of our knowledge, this is the first work that uses trainable noise injection to improve network robustness against adversarial attacks, rather than manually configuring the injected noise level through cross-validation. The extensive results show that our proposed PNI technique effectively improves the robustness against a variety of powerful white-box and black-box attacks such as PGD, C \& W, FGSM, transferable attack and ZOO attack. Last but not the least, PNI method \textbf{improves both clean- and perturbed-data accuracy} in comparison to the state-of-the-art defense methods, which outperforms current unbroken PGD defense by 1.1 \% and 6.8 \% on clean test data and perturbed test data respectively using Resnet-20 architecture.

\end{abstract}

\section{Introduction}

\begin{figure}[t]
  \centering
   \includegraphics[width=0.4\textwidth,height=0.3\textwidth]{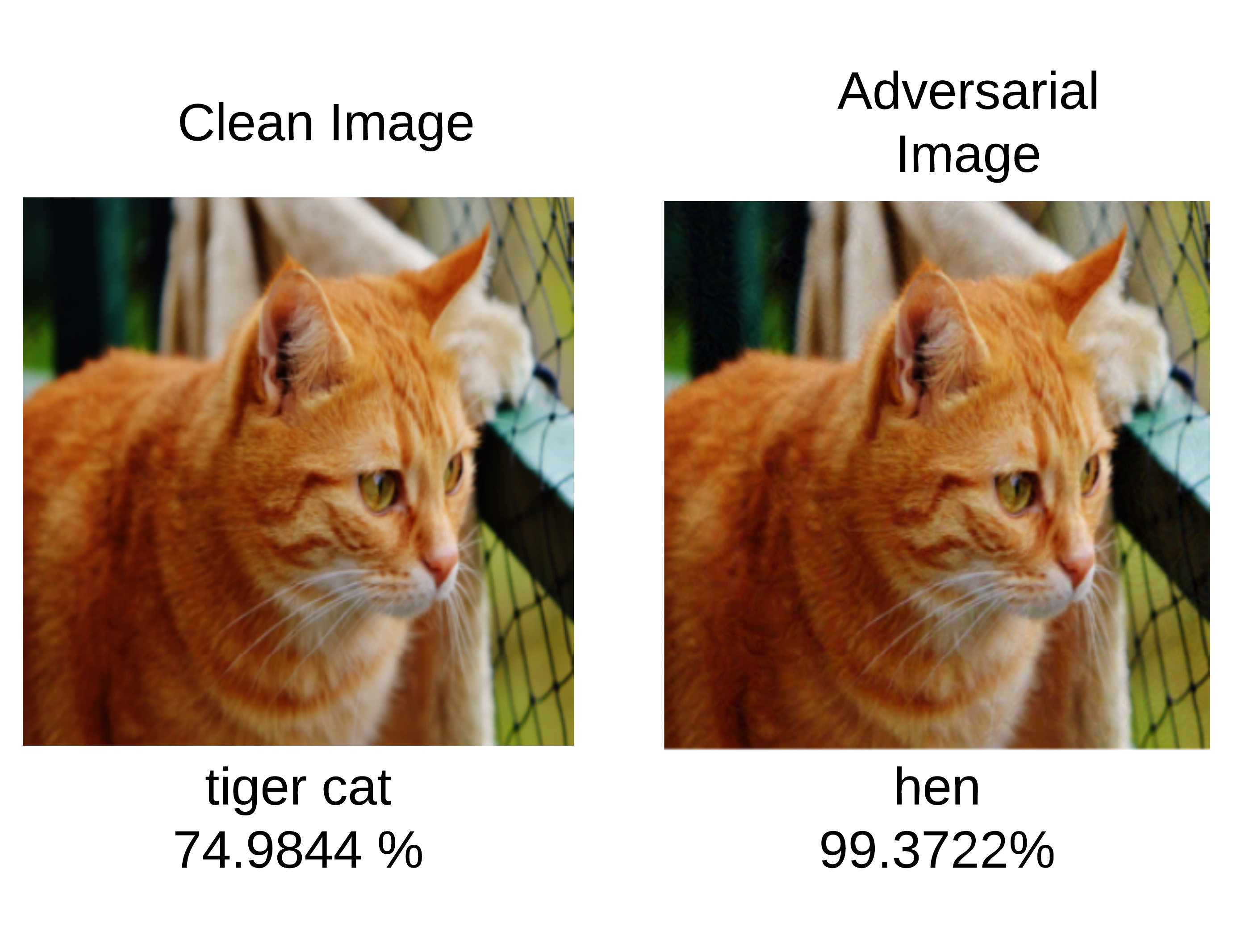}
   \caption{Adversarial attack miss-classifying a cat image to a hen with higher confidence. }
   \label{adv}

\end{figure}

Deep Neural Networks (DNNs) have achieved great success in a variety of applications, including but not limited to image classification~\cite{krizhevsky2012imagenet},  speech recognition \cite{hinton2012deep}, machine translation~\cite{bahdanau2014neural}, and autonomous driving~\cite{chen2015deepdriving}. Despite the remarkable accuracy imrovement \cite{he2015delving}, recent studies~\cite{Szegedy2013IntriguingPO,goodfellow2014explaining,carlini2017towards} have shown that DNNs are vulnerable to adversarial examples. In image classification task, an adversarial example is a natural image intentionally perturbed by visually imperceptible variation, but can cause drastic classification accuracy degradation. \cref{adv} provides an illustration of adversarial example and its original counterpart. 
In addition to image classification, attacks to other DNN-powered tasks have also been actively investigated, such as visual question answering \cite{xu2017can,akhtar2018threat}, image captioning \cite{chen2017show}, semantic segmentation \cite{metzen2017universal,akhtar2018threat} and etc \cite{cheng2018seq2sick,carlini2018audio, sun2018identify}.


There has been a cohort of works on generating adversarial attacks and developing corresponding defense methods. The adversarial attacks can be categorized as \textit{white-box attack} and \textit{black-box attack} based on the attacker's knowledge to the target model. For white-box attack \cite{Szegedy2013IntriguingPO,carlini2017towards}, the adversary has full access to the network architecture and parameters. Whereas, only the input and output to the network can be externally accessed by the black-box attacks \cite{liu2016delving,papernot2016transferability,chen2017zoo}. White-box attack can often achieve high success rates for various applications \cite{Szegedy2013IntriguingPO,kurakin2016adversarial,kos2017delving,papernot2016limitations,moosavi2016deepfool,athalye2018obfuscated,chen2017zoo,carlini2017towards}.

Recently, different works \cite{bietti2018regularization} have viewed the problem of adversarial examples from an unified perspective of model robustness and regularization. Conventional regularization mainly serves the purpose of reducing the generation error, thus preventing model from overfiting the training set.
Traditional regularization methods have been effective in neural network training. For example, dropout \cite{JMLR:v15:srivastava14a}, Batch Normalization (BN) \cite{ioffe2015batch} and quantization \cite{zhou2016dorefa,courbariaux2015binaryconnect} all serve the purpose of model regularization. However, BN is specifically effective in convolution networks and dropout is applicable for fully connected network.
Hinton discusses that adding Gaussian noise into the model (input, weight and activation) during training performs as a regularizer in his lecture note\footnote{\url{https://www.cs.toronto.edu/~tijmen/csc321/slides/lecture_slides_lec9.pdf}} and dropout work \cite{JMLR:v15:srivastava14a}.

It is evident that a more general model regularization method specifically directed for improving neural network robustness can serve the purpose of defending adversarial example more effectively. Recently, different works have implemented noise regularization method both during training and inference phases \cite{liu2017towards,lecuyer2018certified,yoshida2017spectral,bietti2018regularization}. In this work, we propose to improve neural network robustness against adversarial attack by regularizing the training process through adding noise during the training phase. We believe adversarial defense methodologies that focus on defending the network at the inference only will eventually fall short in the advent of new attack methods. Thus a more general regularized training method can generate robust DNN to defend against a wide range of attacks.

\textbf{Overview of our approach:}
In this work, we propose a novel noise injection method called Parametric Noise Injection (PNI) to improve neural network robustness against adversarial attack. It has the flexibility to inject trainable noise at the input (to whole network), activation and weights during both training and inference. The proposed PNI is embedded with well-known adversarial training, where Gaussian Noise with trainable parameters could adjust injected noise level at each neural network layer. 
In this work, we conduct a wide range of white-box and black-box adversarial attack experiments to demonstrate the effectiveness of our proposed PNI method accross different popular DNN architectures. Our simulation shows accuracy improvement for both clean data accuracy and under attack accuracy. PNI achieves a 1.1 \% improvement on the clean test data on Resnet-20 compared to Vanilla Resnet-20 with adversarial training. Along with the improvement on clean test data, our defense shows 6.8\% improvement on the test accuracy under PGD wite box attack. Additionally, Our result shows improved robustness under FGSM, C \& W attack and various black-box attack.

\section{Related works}

\subsection{Adversarial Attack}
\label{sec:adversarial_attck_intro}
Recently, various powerful adversarial attack methods have been proposed to totally fool a trained deep neural network through introducing barely visible perturbation upon input data. Several state-of-the-art white-box (i.e., PGD \cite{madry2018towards}, FGSM \cite{goodfellow2014explaining} and C\&W \cite{carlini2017towards}) and black-box (i.e., Substitute \cite{papernot2017practical} and ZOO \cite{chen2017zoo}) adversarial attack method are briefly introduced as follows.

\paragraph{FGSM Attack:}
Fast Gradient Sign Method (FGSM) \cite{Szegedy2013IntriguingPO} is a single-step efficient adversarial attack method, which alters each element $x$ of nature sample $\bm{x}$ along the direction of its gradient w.r.t the loss function $\partial \mathcal{L}/\partial x$. The generation of adversarial example $\bm{\hat{x}}$ can be described as:
\begin{equation}
\label{eqt:fgsm}
    \bm{\hat{x}} = \bm{x} + \epsilon \cdot sgn\big(\nabla_{\bm{x}}\hat{\mathcal{L}}(g(\bm{x};\bm{\theta}),t)\big)
\end{equation}
where the attack is followed by a clipping operation to ensure the $\hat{x}\in [0,1]$. The attack strength is determined by the perturbation constraint $\epsilon$.

\paragraph{PGD Attack:} Projected Gradient Descent (PGD) \cite{madry2018towards} is the multi-step variant of FGSM, which is one of the strongest $L^\infty$ adversarial example generation algorithm. With $\bm{\hat{x}}^{1} = \bm{x}$ as the initialization, the iterative update of perturbed data $\bm{\hat{x}}$ can be expressed as:
\begin{equation}
\label{eqt:pgd}
    \bm{\hat{x}}^{t+1}=\Pi_{P_\epsilon(\bm{x})} \Big( \bm{\hat{x}}^t + a \cdot sgn\big(\nabla_{\bm{x}}\hat{\mathcal{L}}(g(\bm{\hat{x}}^t;\bm{\theta}),t)\big)\Big)
\end{equation}
where $P_\epsilon(\bm{x})$ is the projection space which is bounded by $\bm{x}\pm\epsilon$, $t$ is the step index up to $N_\textup{step}$, and $a$ is the step size. Madry et al. \cite{madry2018towards} proposed that PGD is a universal adversary among all the first-order adversaries (i.e., attacks only rely on first-order information).

\paragraph{C \& W Attack:}
In C \& W attack method, Carlini and Wagner \cite{carlini2017towards} consider the generation of adversarial example as an optimization problem, which optimize the $L^p$-norm of distance metric $\delta$ w.r.t the given input data $\bm{x}$, which can be described as:
\begin{equation}
\label{eqt:CW_Lp_norm}
    ||\bm{\delta}||_p = \Big( \sum_{i=1}^n |\delta_i|^p \Big)^{1/p} ;\quad \delta_i = \hat{x}_i-x_i
\end{equation}
\begin{equation}
\label{eqt:CW_minimize}
    \textup{minimize}~||\bm{\delta}||_p + c\cdot \hat{\mathcal{L}}(\bm{x}+\bm{\delta})\quad s.t.\quad \bm{x}+\bm{\delta}\in [0,1]^n
\end{equation}
where $\bm{\delta}$ is taken as the perturbation added upon the input data, and a proper loss function $\hat{\mathcal{L}}$ is chosen in \cite{carlini2017towards} to to solve the optimization problem via gradient descent method. $c$ is a constant set by attacker. In this work, we use $L^2$-norm based C\&W attack and take $||\bm{\delta}||_{p=2}$ as the evaluation metric to measure the network's robustness, where a higher value of $||\bm{\delta}||_{p=2}$ indicates a more robust network or potential failure of the attack. 

\paragraph{Black-box Attacks:}
The most popular black-box attack is conducted using a substitute model \cite{papernot2017practical}, where the attacker trains a substitute model to mimic the functionality of target model, then use the adversarial example generated from the substitute model to attack target model. In this work, we specifically investigate the transferable adversarial attack \cite{liu2016delving}, which is a variant of substitute model attack. In transferable adversarial attack, the adversarial example is generated from one source model to attack another target model. The source model and target can own the absolutely different structure but trained on the identical dataset. Moreover, Zero-th Order Optimization (ZOO) attack \cite{chen2017zoo} is also considered. Rather than training a substitute model, it directly approximates the gradient of target model just based on the input data and output scores using stochastic gradient coordinate.

\subsection{Adversarial Defenses:}
\label{sub:related}
Improving network robustness by training the model with adversarial examples \cite{Szegedy2013IntriguingPO,madry2018towards} is the most popular defense approach now-a-days. Most of later works have followed this path to supplement their defense with adversarial training \cite{buckman2018thermometer,samangouei2018defensegan}. The first step in adversarial training is to choose an attack model to generate adversarial examples. Adopting Projected Gradient Descent (PGD) based attack model to adversarial training is becoming popular since it can generate universal adversarial examples among the first order approaches \cite{madry2018towards}. Additionally, among many recent defense methods, only PGD based adversarial training can sustain state-of-the-art accuracy under attacks \cite{carlini2017towards,Szegedy2013IntriguingPO,athalye2018obfuscated}. The reported DNN accuracy in CIFAR10 dataset remains a major success to defend very strong adversarial attacks \cite{athalye2018obfuscated}.

Recent works have merged the concept of improving model robustness through regularization to defend adversarial examples. Among them, an unified perspective of regularization and robustness was presented by \cite{bietti2018regularization}. Again, randomly pruning some activation during the inference \cite{Santhanam2018DefendingAA} or randomizing the input layer \cite{xie2018mitigating} serve the purpose of injecting randomness to somehow prevent the attacker from accessing the gradient. However, these approaches achieve good success against gradient based attacks at the cost of obfuscated gradient \cite{athalye2018obfuscated}.

In order to make the model more robust to adversarial attack, several works have adopted the concept of adding a noise layer just before convolution layer during both training and inference phases \cite{liu2017towards,lecuyer2018connection}. Even though we agree with the core idea of these works as they certainly makes the model more robust, but there are some fundamental advantages of our work compared to theirs. PNI improves the model robustness by regularizing the model while training more effectively. As classical machine learning demonstrated weight noise performs the regularization even better \cite{JMLR:v15:srivastava14a}.  We also show experimentally that particularly adding noise to the weights improves the robustness even more. While these works \cite{liu2017towards,lecuyer2018certified} have chosen level of noise to be injected manually, we propose to inject different level of noise at different layers using trainable parameters. As choosing the level of noise manually for different layers even by validation set is not practically feasible.

\section{Approach}
In this section, we first introduce the proposed Parametric Noise Injection (PNI) function and will investigate the impact of noise injection on input (to the whole DNN), weight and activation.

\subsection{Parametric Noise Injection}


\paragraph{Definition.} The method that we propose to inject noise to different components or locations within DNN can be described as:
\begin{equation}
\label{eqt:PNI_definition}
    \Tilde{v}_i = f_\textup{PNI}(v_i) = v_i + \alpha_i \cdot \eta; \quad \eta \sim \mathcal{N}(0, \sigma^2)
\end{equation}
\begin{equation}
\label{eqt:std_tensor_V}
    \sigma = \sqrt{\dfrac{1}{N}\sum_i (v_i-\mu)}
\end{equation}
where $v_i$ is the element of noise-free tensor $\bm{v}$, and such $\bm{v}$ can be input/weight/inter-layer tensor in this work. $\eta$ is the additive noise term which follows the Gaussian distribution with zero mean and standard deviation $\sigma$, and $\alpha_i$ is the coefficient scales the magnitude of injected noise $\eta$. We adopt the scheme that $\eta$ shares the identical standard deviation of $\bm{v}$ as in \cref{eqt:std_tensor_V}, thus the injected additive noise is correlated to the distribution of $\bm{v}$ and $\alpha$ simultaneously.
Moreover, rather than manually configuring $\alpha_i$ to restrict the noise level, we set $\alpha_i$ as learnable parameter which can be optimized for network robustness improvement. We name such method as \textit{Parametric Noise Injection} (PNI). Considering the over-parameterization and the convergence of training $\alpha_i$, we make the element-wise noise term ($\alpha_i \cdot \eta$) shares the same scaling coefficient across the entire tensor. Assuming we performs the proposed PNI on the weight tensors of convolution/fully-connected layers throughout entire DNN, for each parametric layer there is only one layer-wise noise scaling coefficient to be optimized. We takes such layer-wise configuration as default in this work. 


\paragraph{Optimization}
In this work, we treat the noise scaling coefficient as a model parameter which can be optimized through back-propagation training process. For $f_\textup{PNI}(\cdot)$ configuration which shares the noise scaling coefficient layer-wise, the gradient computation can be described as:
\begin{equation}
    \dfrac{\partial\mathcal{L}}{\partial \alpha} = 
    \sum_i\dfrac{\partial\mathcal{L}}{\partial f_\textup{PNI}(v_i)} \dfrac{\partial f_\textup{PNI}(v_i)} {\partial \alpha}
\end{equation}
where the $\sum_i$ takes the summation over the entire tensor $\bm{v}$, and $\partial \mathcal{L}/\partial f_\textup{PNI}(v_i)$ is the gradient back-propagated from the followed layers. The gradient calculation of the PNI function is:
\begin{equation}
    \dfrac{\partial f_\textup{PNI}(v_i)} {\partial \alpha} = \eta
\end{equation}
It is noteworthy that even though $\eta$ is a Gaussian random variable, each sample of $\eta$ is taken as a constant during the back-propagation. Using the gradient descent optimizer with momentum, the optimization of $\alpha$ at step $j$ can be written as:
\begin{equation}
    V_i^{j} = m \cdot V_i^{j-1} + \dfrac{\partial \mathcal{L}^{j-1}}{\partial \alpha}; \quad 
    \alpha^{j} = \alpha^{j-1} - \epsilon\cdot  V_i^{j} 
\end{equation}
where $m$ is the momentum, $\epsilon$ is the learning rate, and $V$ is the updating velocity. Moreover, since weight decay tends to make the learned noise scaling coefficient converge to zero, there is no weight decay term on the $\alpha$ during the parameter updating in this work.
We set $\alpha = 0.25$ as default initialization.

\paragraph{Robust Optimization.} 
\label{sec:robust_optimization}
We expect to utilize the aforementioned PNI technique to improve the network robustness. However, directly optimizing the noise scaling coefficient normally leads $\alpha$ to converge at a small close-to-zero value, owing to the model optimization tends to over-fit the training dataset (referring to \cref{table:alpha_converge}). 

In order to succeed in adversarial defense, we jointly use the PNI method with robust optimization (a.k.a. \textit{Adversarial Training}) which can boost the inference accuracy for the perturbed data under attack. Given inputs- $\bm{x}$ and target labels- $t$, the adversarial training is to obtain the optimal solution of network parameter $\bm{\theta}$ for the following min-max problem: 
\begin{equation}
\label{eqt:min_max_game}
    \argmin_{\bm{\theta}} \big\{ \argmax_{\bm{x}' \in P_\epsilon(\bm{x})} \mathcal{L}\big( g(\hat{\bm{x}};f_{\textup{PNI}}(\bm{\theta})), t \big) \big\}
\end{equation}
where the inner maximization tends to acquire the perturbed data $\hat{\bm{x}}$, and $P_\epsilon(\bm{x})$ is the input data perturb set constrained by $\epsilon$. While the outer minimization is optimized through gradient descent method as regular network training. $L^\infty$ PGD attack \cite{madry2018towards} is adopted as the default inner maximization solver (i.e., generating $\hat{\bm{x}}$). Note that, in order to prevent the label leaking during adversarial training, the perturbed data $\hat{\bm{x}}$ is generated through taking the predicted result of $\bm{x}$ as the label (i.e. $t$ in \cref{eqt:pgd}).

Moreover, in order to balance the clean data accuracy and perturbed data accuracy for practical application, rather than performing the outer minimization solely on the loss of perturbed data as in \cref{eqt:min_max_game}, we minimize the ensemble loss $\mathcal{L}'$ which is the weighted sum of losses for clean- and perturbed-data. The ensemble loss is described as:
\begin{equation}
\label{eqt:ensemble_loss}
\resizebox{.42\textwidth}{!}{
   $\mathcal{L}'=w_c \cdot \mathcal{L}(g(\bm{x};f_{\textup{PNI}}(\bm{\theta})), t) + w_a\cdot \mathcal{L}(g(\hat{\bm{x}};f_{\textup{PNI}}(\bm{\theta})),t)$
   }
\end{equation}
where $w_c$ and $w_a$ are the weights for clean data loss and adversarial data loss. $w_c=w_a=0.5$ is the default configuration in this work. Optimizing the ensemble loss $\mathcal{L}'$ with gradient decent method leads to successful training of $f_{\textup{PNI}}(\bm{\theta})$ for both the model's inherent parameter (e.g. weight, bias) and the add-on noise scaling coefficient $\alpha$ from PNI.

\section{Experiments}

\subsection{Experiment setup}
\paragraph{Datasets and network architectures.} 
The CIFAR-10 \cite{krizhevsky2009learning} dataset is composed of 50K training samples and 10K test samples of 32$\times$32 color image. For CIFAR-10, the classical Residual Networks \cite{he2016deep} (ResNet-20/32/44/56) architecture are used, and ResNet-20 is taken as the baseline for most of the comparative experiments and ablation studies. A redundant network ResNet-18 is also used to report the performance for CIFAR-10, since large network capacity is helpful for adversarial defense. Moreover, rather than including the input normalization within the data augmentation, we place a non-trainable data normalization layer in front of the DNN to perform the identical function, thus attacker can directly add the perturbation on the nature image. Note that, since both PNI and PGD attack \cite{madry2018towards} include randomness, we report the accuracy in the format of mean$\pm$std\% with 5 trials to alleviate error.

\paragraph{Adversarial attacks.} 
To evaluate the performance of our proposed PNI technique, we employ multiple powerful white-box and black-box attacks as introduced in \cref{sec:adversarial_attck_intro}. For PGD attack on MNIST and CIFAR-10, $\epsilon$ is set to 0.3/1 and 8/255, and $N_\textup{step}$ is set to 40 and 7 respectively. FGSM attack adopt the same $\epsilon$ setup as PGD. The attack configurations of PGD and FGSM are identical as the setup in \cite{buckman2018thermometer, madry2018towards}.
For C\&W attack, we set the constant $c$ as 0.01. ADAM \cite{kingma2014adam} is used to optimize the \cref{eqt:CW_minimize} with learning rate as $5e^{-4}$. We choose 0 for the confidence coefficient $k$, which is defined in $\hat{\mathcal{L}}$ used by C\&W $L^2$ attack in \cite{carlini2017towards}. The binary search steps for the attack is 9, while number of iteration to perform the gradient descent is 10. Moreover, We also conduct the PNI defense against several state-of-the-art black-box attacks (i.e. substitute \cite{papernot2017practical}, ZOO \cite{chen2017zoo} and transferable \cite{liu2016delving} attack) in a \cref{sec:black_box_defense} to examine the robustness improvement resulted from the proposed PNI technique.

\paragraph{Competing methods for adversarial defense.} As far as we know, the adversarial training with PGD \cite{madry2018towards} is the only unbroken defense method \cite{athalye2018obfuscated}, which is labeled as \textit{vanilla adversarial training} and taken as the baseline in this work. Beyond that, several recent works also utilize similar concept as ours in their defense method are discussed as well, including certified robustness \cite{lecuyer2018certified} and random self-ensemble \cite{liu2017towards}.

\subsection{PNI for adversarial attacks}
\subsubsection{PNI against white-box attacks}

\begin{table}[t]
\centering
\caption{\textbf{Convergence of PNI}: ResNet-20 with Layerwise weight PNI on CIFAR-10 dataset. (Top) The converged layer-wise noise scaling coefficient $\alpha$ under various training scheme. (Bottom) Test accuracy for clean- and perturbed-data under PGD and FGSM attack.}
\label{table:alpha_converge}
\resizebox{0.45\textwidth}{!}{%
\begin{tabular}{@{}cccc@{}}
\toprule
\begin{tabular}[c]{@{}c@{}}Layer\\ Index\end{tabular} & \begin{tabular}[c]{@{}c@{}}Vanilla\\ Traning\end{tabular} & \begin{tabular}[c]{@{}c@{}}PNI-W+Adv. Train.\\ (without PNI in \\ $\hat{\bm{x}}$ generation)\end{tabular} & \begin{tabular}[c]{@{}c@{}}PNI-W+Adv. Train.\\ (with PNI in \\ $\hat{\bm{x}}$ generation)\end{tabular} \\ \midrule
Conv0 & 0.003 & 0.004 & \textbf{0.146} \\ \midrule
Conv1.0 & 0.002 & 0.005 & \textbf{0.081} \\
Conv1.1 & 0.004 & 0.004 & \textbf{0.049} \\
Conv1.2 & 0.002 & 0.001 & \textbf{0.097} \\
Conv1.3 & 0.004 & 5.856 & \textbf{0.771} \\
Conv1.4 & 0.005 & 0.005 & 0.004 \\
Conv1.5 & 0.002 & 0.001 & 0.006 \\ \midrule
Conv2.0 & 0.004 & 0.000 & 0.006 \\
Conv2.1 & 0.006 & 0.003 & 0.004\\
Conv2.2 & 0.004 & 0.003 & 0.030 \\
Conv2.3 & 0.001 & 0.006 & 0.003\\
Conv2.4 & 0.003 & 0.001 & 0.033 \\
Conv2.5 & 0.002 & 0.001 & 0.023 \\ \midrule
Conv3.0 & 0.007 & 0.001 & 0.008 \\
Conv3.1 & 0.003 & 0.001 & 0.006 \\
Conv3.2 & 0.007 & 0.002 & 0.001 \\
Conv3.3 & 0.006 & 0.001 & 0.002 \\
Conv3.4 & 0.009 & 0.002 & 0.001\\
Conv3.5 & 0.005 & 0.000 & 0.001\\ \midrule
FC & 0.002  & 0.002 & 0.001 \\ \midrule 
\midrule
Clean & 92.11\% & 71.00\% & 84.89$\pm$0.11\% \\
PGD & 0.00$\pm$0.00\% & 18.11\% & 45.94$\pm$0.11\% \\
FGSM & 14.08\% & 26.34\% & 54.48$\pm$0.44\% \\
\bottomrule
\end{tabular}
}
\end{table}

\begin{figure}[t]
	\centering
	\begin{tabular}{c}
	\includegraphics
	[width=0.95\linewidth]{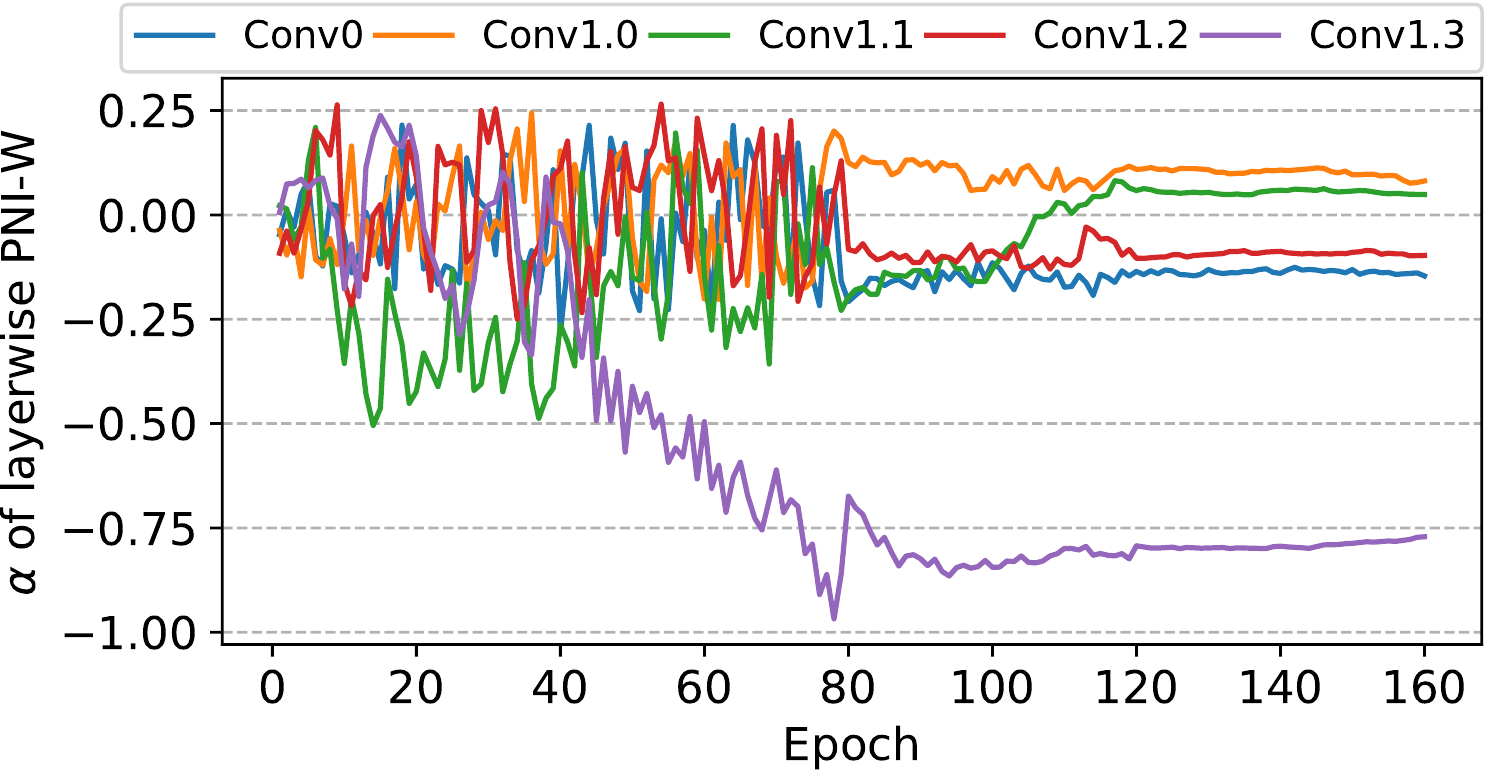}\\
	\end{tabular}
	\caption{The evolution curve of trainable noise scaling coefficient $\alpha$ for layerwise PNI on weight (PNI-W). Only front 5 layers (bold in \cref{table:alpha_converge}) of ResNet-20 \cite{he2016deep} are shown. The learning rate of SGD optimizer is reduced at 80 and 120 epoch.}
	\label{fig:alpha_converge}

\end{figure}

\paragraph{Optimization method of PNI}
As the aforementioned discussion in \cref{sec:robust_optimization}, the noise scaling coefficient will not be properly trained without utilizing the adversarial training (i.e., solving the min-max problem). We conduct the experiments for training the layer-wise PNI on weight (PNI-W) of ResNet-20, to compare the convergence of trained noise. As tabulated in \cref{table:alpha_converge}, simply performing the vanilla training using momentum SGD optimizer totally fails the adversarial defense, where the noise scaling coefficients $\alpha$ are converged to the negligible values. On the contrary, with the aid of adversary training (i.e., optimization of \cref{eqt:ensemble_loss}), convolution layers in the network's front-end has obtained relatively large $\alpha$ which are the bold values in \cref{table:alpha_converge}, and the corresponding evolution curve are shown in \cref{fig:alpha_converge}.

Since the PGD attack \cite{madry2018towards} is taken as the inner maximization solver, the generation of adversarial example $\hat{\bm{x}}$ in \cref{eqt:pgd} is reformatted as:
\begin{equation}
\label{eqt:pgd_with_PNI}
\resizebox{.42\textwidth}{!}{
    $\bm{\hat{x}}^{t+1}=\Pi_{P_\epsilon(\bm{x})} \Big( \bm{\hat{x}}^t + a \cdot sgn\big(\nabla_{\bm{x}}\hat{\mathcal{L}}(g(\bm{\hat{x}}^t;f_{\textup{PNI}}(\bm{\theta})),t)\big)\Big)$
    }
\end{equation}
where the difference between \cref{eqt:pgd} and \cref{eqt:pgd_with_PNI} is with/without PNI within $\hat{\bm{x}}$ generation. It is noteworthy that, keeping the noise term in the model for both adversarial example generation (\cref{eqt:pgd_with_PNI}) and model parameter update is the critical factor for the PNI optimization with adversarial training, since the optimization of $\alpha$ is also a min-max game. Increasing noise level enhances the defense strength, but hampers the network inference accuracy for natural clean image. Lowering $\alpha$, however, makes the network vulnerable to adversarial attack. As listed in \cref{table:alpha_converge}, without PNI-W in $\hat{\bm{x}}$ generation indeed leads to the failure of PNI optimization, and the large value ($\alpha=5.856$ in \cref{table:alpha_converge}) is not converged due to the probable gradient explosion.



\begin{table*}[t]
\centering
\caption{\textbf{Effect of PNI location}: The ResNet-20 \cite{he2016deep} clean- and perturbed-data (under PGD and FGSM attack) accuracy (mean$\pm$std\%) on CIFAR-10 test-set, with PNI technique on different network location. Baseline is the ResNet-20 with vanilla adversarial training, and all the PNI combinations are optimized through adversarial training by default.}
\label{table:PNI_schemes}
\resizebox{0.7\textwidth}{!}{%
\begin{tabular}{@{}ccccccc@{}}
\toprule
 & \multicolumn{3}{c}{Test with PNI} & \multicolumn{3}{c}{Test without PNI} \\ 
 \cmidrule(l){2-7} 
 & Clean & PGD & FGSM & Clean & PGD & FGSM \\ \midrule
Vanilla adv. train \cite{madry2018towards}& - & - & - & 83.84 & 39.14$\pm$0.05 & 46.55 \\ \midrule
PNI-W & 84.89$\pm$0.11 & \textbf{45.94$\pm$0.11} & \textbf{54.48$\pm$0.44} & 85.48 & 31.45$\pm$0.07 & 42.55 \\
PNI-I & 85.10$\pm$0.08 & 43.25$\pm$0.16 & 50.78$\pm$0.16 & 84.82 & 34.87$\pm$0.05 & 44.07 \\
PNI-A-a & 85.22$\pm$0.18 & 43.83$\pm$0.10 & 51.41$\pm$0.08 & 85.20 & 33.93$\pm$0.05 & 44.32 \\ 
PNI-A-b & 84.66$\pm$0.16 & 43.63$\pm$0.20 & 51.26$\pm$0.09 & 83.97 & 33.53$\pm$0.05 & 43.37 \\ \midrule
PNI-W+A-a & 85.12$\pm$0.10 & 43.57$\pm$0.12 & 51.15$\pm$0.21 & 84.88 & 33.23$\pm$0.05 & 43.59 \\ 
PNI-W+A-b & 84.33$\pm$0.11 & 43.80$\pm$0.19 & 51.14$\pm$0.07 & 84.42 & 33.30$\pm$0.05 & 43.43 \\ \bottomrule
\end{tabular}%
}
\end{table*}

\begin{table*}[t]
\centering
\caption{\textbf{Effect of network depth and width}: The clean- and perturbed-data (under PGD and FGSM attack) accuracy (mean$\pm$std\%) on CIFAR-10 test-set, utilizing different robust optimization configurations. For network depth, the classical ResNet-20/32/44/56 with increasing depth is reported. For network width, the ResNet-20 (1$\times$) is adopted as the baseline, then we compare the wide ResNet-20 with the input and output channel scaled by 1.5$\times$/2$\times$/4$\times$. Capacity denotes the number of trainable parameters in the model.}
\label{table:PNI-w_comparison}
\resizebox{\textwidth}{!}{%
\begin{tabular}{@{}cccccccccccccc@{}}
\toprule
 &  & \multicolumn{3}{c}{No defense} & \multicolumn{3}{c}{Vanilla adv. train} & \multicolumn{3}{c}{\begin{tabular}[c]{@{}c@{}}PNI-W+adv. train\\ (Test with PNI)\end{tabular}} & \multicolumn{3}{c}{\begin{tabular}[c]{@{}c@{}}PNI-W+adv. train\\ (Test without PNI)\end{tabular}} \\
 \cmidrule(l){3-14} 
Model & Capacity & Clean & PGD & FGSM & Clean & PGD & FGSM & Clean & PGD & FGSM & Clean & PGD & FGSM \\ \midrule
Net20  & 269,722 & 92.1 & 0.0$\pm$0.0 & 14.1 & 83.8 & 39.1$\pm$0.1 & 46.6 & 84.9$\pm$0.1 & 45.9$\pm$0.1 & 54.5$\pm$0.4 & 85.5 & 31.6$\pm$0.1 & 42.6 \\ \midrule
Net32  & 464,154 & 92.8 & 0.0$\pm$0.0 & 17.8 & 85.6 & 42.1$\pm$0.0 & 50.3 & 85.9$\pm$0.1 & 43.5$\pm$0.3 & 51.5$\pm$0.1 & 86.4 & 35.3$\pm$0.1 & 45.5 \\
Net44  & 658,586 & 93.1 & 0.0$\pm$0.0 & 23.9 & 85.9 & 40.8$\pm$0.1 & 48.2 & 84.7$\pm$0.2 & 48.5$\pm$0.2 & 55.8$\pm$0.1 & 86.0 & 39.6$\pm$0.1 & 49.9 \\
Net56  & 853,018 & 93.3 & 0.0$\pm$0.0 & 24.2 & 86.5 & 40.1$\pm$0.1 & 48.8 & 86.8$\pm$0.2 & 46.3$\pm$0.3 & 53.9$\pm$0.1 & 87.3 & 41.6$\pm$0.1 & 51.1 \\ \midrule
Net20(1.5$\times$) & 605,026 & 93.5 & 0.0$\pm$0.0 & 15.9 & 85.8 & 42.0$\pm$0.0 & 49.6 & 86.0$\pm$0.1 & 46.7$\pm$0.2 & 54.5$\pm$0.2 & 87.0 & 38.4$\pm$0.1 & 49.1 \\ 
Net20(2$\times$)  & 1,073,962 & 94.0 & 0.0$\pm$0.0 & 13.0 & 86.3 & 43.1$\pm$0.1 & 52.6 & 86.2$\pm$0.1 & 46.1$\pm$0.2 & 54.6$\pm$0.2 & 86.8 & 39.1$\pm$0.0 & 50.3 \\ 
Net20(4$\times$)  & 4,286,026 & 94.0 & 0.0$\pm$0.0 & 14.2 & 87.5 & 46.1$\pm$0.1 & 54.1 & 87.7$\pm$0.1 & 49.1$\pm$0.3 & 57.0$\pm$0.2 & 88.1 & 43.8$\pm$0.1 & 54.2 \\ 
\bottomrule
\end{tabular}%
}
\end{table*}

\paragraph{Effect of PNI on weight, activation and input.}
In this work, even though the scheme of injecting noise on the weight (PNI-W) is taken as the default PNI setup, more results about PNI on activation (PNI-A-a/b), input (PNI-I) and hybrid-mode (e.g. PNI-W+A) are provided in \cref{table:PNI_schemes} for a comprehensive study.
PNI-A-a/PNI-A-b denotes injecting noise on the output/input tensor of the convolution/fully-connected layer respectively. Moreover, PNI-A-b scheme intrinsically includes the PNI-I, since PNI-I is applying the noise on the input tensor of first layer.
Note that, all models with PNI variants are jointly trained with PGD-based adversarial training \cite{madry2018towards} as discussed above. Then, with the same trained model, we report the accuracy with/without the trained noise term (left/right in \cref{table:PNI_schemes}) during the test phase.
As shown in \cref{table:PNI_schemes}, with the noise term enabled during test phase, PNI-W on ResNet-20 gives the best performance to defend PGD and FGSM attack, in comparison to PNI on other locations. Although it is elusive to fully understand the mechanism that PNI-W outperforms other counterparts, the intuition is that PNI-W is the generalization of PNI-A in each connection instead of each output unit, similar as relation between the regularization technique DropConnect \cite{wan2013regularization} and Dropout \cite{JMLR:v15:srivastava14a}. 

Furthermore, we also observe that disabling PNI during test phase leads to significant accuracy drop for defending PGD and FGSM attack, while the clean-data accuracy maintains the same level as PNI enabled. Such observation raises two concerns about our PNI techniques: 1) Does the improvement of clean-/perturbed-data accuracy with PNI mainly comes from the attack strength reduction caused by the randomness (potential gradient obfuscation \cite{athalye2018obfuscated})? 2) Is PNI just an negligible trick or it performs the model regularization to construct a more robust model? Our answers to both questions are negative, where the explanations are elaborated under \cref{sec:core_problem_answer}.

\paragraph{Effect of network capacity.}
In order to investigate the relation between network capacity (i.e., number of trainable parameters) and robustness improvement by PNI, we examine various network architectures in terms of both depth and width. For different network depths, experiments on ResNet 20/32/44/56 \cite{he2016deep} are conducted under vanilla adversarial training \cite{madry2018towards} and our proposed PNI robust optimization method. For different network widths, we adopt the original ResNet-20 as baseline and expand its input\&output channel of each layer by 1.5$\times$/2$\times$/4$\times$ respectively. Same as \cref{table:PNI_schemes}, we report clean- and perturbed-data accuracy with/without PNI term during the test phase. The results in \cref{table:PNI-w_comparison} indicates that increasing the model's capacity indeed improves network robustness against white-box adversarial attacks, and our proposed PNI outperforms vanilla adversary training in terms of both clean-data accuracy and perturbed data accuracy for PGD and FGSM attack. Such observation demonstrates that the perturbed-data accuracy improvement does not come from trading off clean-data accuracy as reported in \cite{buckman2018thermometer,anonymous2019l2-nonexpansive}. Through increasing the network capacity, the drop perturbed-data accuracy, when disabling the PNI noise term during test phase, also becomes less significant. Although both adversarial training and PNI techniques perform regularization, the network structure still needs careful construction to prevent the over-fitting resulted from over-parameterization.


\paragraph{Robustness evaluation with C\&W attack.}

Improved robustness does not necessarily mean improving the test data accuracy against any particular attack method. Typically $L_2$ norm based C \& W attack \cite{carlini2017towards} should reach 100 \% success rate against any defense. Thus average $L_2$ norm required to fool the network gives more insight about a network's robustness in general \cite{carlini2017towards}. The result presented in \cref{table:CW_defense} represents the overall performance of our model against C \& W attack. Our method of training the noise parameter becomes more effective for more redundant network. We demonstrate this phenomena by performing comparison study between Resnet-20 and Resnet-18 architecture. Clearly Resnet-18 shows the improvement in robustness from Vanilla adv. training much more than Resnet-20 against C \& W attack.

\begin{table}[t]
\centering
\caption{C \& W attack $L_2$ norm comparison}
\label{table:CW_defense}
\resizebox{0.47\textwidth}{!}{%
\begin{tabular}{@{}ccccc@{}}
\toprule
 &  & \multicolumn{3}{c}{CW L2-norm} \\ \cmidrule(l){3-5} 
Model & capacity & No defense & Vanilla adv. train & PNI-W \\ \midrule
ResNet-20 (4x) & 4,286,026 & 0.12 & 1.97 & 1.97 \\
ResNet-18 & 11,173,962 & 0.12 & 2.39 & 2.63 \\ \bottomrule
\end{tabular}%
}
\end{table}

\subsubsection{PNI against black-box attack} 
\label{sec:black_box_defense}
In this section, we test our proposed PNI technique against transferable adversarial attack \cite{liu2016delving} and ZOO attack. 
Following the transferable adversarial attack \cite{liu2016delving}, two trained neural network are taken as the source model ($S$) and target model ($T$). The adversarial examples $\hat{\bm{x}}_s$ is generated from the source model then attack the target model using $\hat{\bm{x}}_s$, which is denoted as $S\Rightarrow T$. We take ResNet-18 on CIFAR-10 as an example. We train two ResNet-18 model (model-A and B) on CIFAR-10 dataset to attack each other, where model-A is optimized through vanilla adversarial training, while model-B is trained using our proposed PNI variants (i.e., PNI-W/A-a/W+A-a) robust optimization method. \cref{table:PNI_black_box} shows almost equal perturbed-data accuracy for A $\Rightarrow$ B and B $\Rightarrow$ A under various PNI scenarios, which indicates that our PNI technique does not reduce the attack strength.

\begin{table}[ht]
\centering
\caption{\textbf{PNI against black-box attacks}: On CIFAR-10 test-set, (Left) perturbed-data accuracy under transferable PGD attack, and (Right) the attack success rate for ZOO attack. Model-A is a ResNet-18 trained by vanilla adversarial training, and Model-B is a ResNet-18 trained by PNI-W/A-a/W+A-a with adversarial training.}
\label{table:PNI_black_box}
\resizebox{0.45\textwidth}{!}{%
\begin{tabular}{@{}cccc@{}}
\toprule
 & \multicolumn{2}{c}{Transferable attack} & ZOO attack \\ \cmidrule(l){2-4}
Train. scheme of B & A $\Rightarrow$ B & B $\Rightarrow$ A & success rate \\ \midrule
PNI-W & 75.13$\pm$0.17 &  75.23$\pm$0.18 & 57.72 \\
PNI-A-a & 74.67$\pm$0.11 & 75.86$\pm$0.13 & 69.61 \\
PNI-W+A-a & 75.14$\pm$0.10 & 74.92$\pm$0.13 & 50.00 \\ \bottomrule
\end{tabular}
}
\end{table}

For ZOO attack\cite{chen2017zoo}, we test our defense on 200 randomly selected test samples for un-targeted attack. The Attack success rate denotes the percentage of test sample change their classification to a wrong class after attack. ZOO attack success rate for vanilla Resnet-18 with adversarial training is close to ~80 \%. The robustness of PNI is more evident from \cref{table:PNI_black_box} as the attack success rate drops significantly for PNI-W+A-a and PNI-W. However, PNI-A-a fails to resist ZOO attack even though it still maintains a lower success rate than baseline. The failure of PNI-A-a shows that just adding noise in-front of the activation does not necessarily achieves the desired robustness as claimed by some of the previous defenses \cite{lecuyer2018certified,liu2017towards}.

\subsubsection{Comparison to competing methods}

As discussed in \cref{sub:related}, a large number of adversarial defense works have been proposed recently, however most of them are already broken by stronger attacks proposed in \cite{DBLP:journals/corr/abs-1804-03286, athalye2018obfuscated}. As a result, in this work we choose to compare with the most effective one till date - PGD based adversarial training \cite{madry2018towards}. Additionally, we compare with other randomness-based works \cite{liu2017towards,lecuyer2018certified} in \cref{tb:comparision} for examining the effectiveness of PNI.


\begin{table}[ht]
\centering
\caption{Comparison of state-of-the-art adversarial defense methods with clean- and perturbed-data accuracy on CIFAR-10 under PGD attack.}
\label{tb:comparision}
\resizebox{0.42\textwidth}{!}{%
\begin{tabular}{@{}cccc@{}}
\toprule
Defense method & model & Clean & PGD \\ \midrule
PGD adv. train \cite{madry2018towards} & ResNet-20 (4$\times$) & 87 & 46.1 \\
DP \cite{lecuyer2018certified} & \begin{tabular}[c]{@{}c@{}}28-10 Wide ResNet\\ (L=0.1)\end{tabular} & 87.0 & 25 \\
RSE \cite{liu2017towards} & ResNext & 87.5 & 40 \\
PNI-W (this work)& ResNet-20 (4$\times$) & 87.7 & 49.1 \\ \bottomrule
\end{tabular}%
}
\end{table}

Previous defense works \cite{anonymous2019l2-nonexpansive,buckman2018thermometer} have shown a trade-off between clean-data accuracy and perturbed-data accuracy, where the perturbed-data accuracy improvement normally at the cost of lowering the clean-data accuracy. It is worthy to highlight that \textbf{our proposed PNI improves both clean- and perturbed data accuracy under white-box attack, in comparison to PGD-based adversarial training} \cite{madry2018towards}. Differential Privacy (DP) \cite{lecuyer2018certified} is a similar method of utilizing noise injection at various locations within the network. Although their defense guarantees a certified defense it does not perform well against $L^\infty$-norm based attack (e.g., PGD and FGSM). In order to achieve a higher level of certified defense, DP significantly sacrifices the clean-data accuracy as well. Another randomness-based approach is Random Self-ensemble (RSE) \cite{liu2017towards}, which inserts noise-layer before all the convolution layer. Even though their defense performs well against C \& W attack but poor against strong PGD attack. In our black-box attack simulation \cref{table:PNI_black_box} we demonstrate that adding activation noise may not be as effective as weight noise. Beyond that, both DP and RSE manually configure the noise level which is extremely difficult to find the optimal setup. Whereas, in our proposed PNI method, the noise level is determined by a trainable layer-wise noise scaling coefficient and distribution of noise injected location.

\section{Discussion}
\label{sec:core_problem_answer}

The defense performance improvement led by our proposed PNI does not come from the stochastic gradients. The stochastic gradient is considered to incorrectly approximate the true gradient based on a single sample. We try to show that PNI is not relying on the gradient obfuscation from two perspectives: 1) Our proposed PNI method passes each inspection item proposed by \cite{athalye2018obfuscated} to identify gradient obfuscation. 2) Under PGD attack, through increasing the attack steps, our PNI robust optimization method still outperforms vanilla adversarial training (certified as non-obfuscated gradients in \cite{athalye2018obfuscated}).

\begin{table}[ht]
\centering
\caption{Checklist of examining the characteristic behaviors caused by obfuscated and masked gradient \cite{athalye2018obfuscated} for PNI.}
\label{tabel:checklist_gradient_obfuscation}
\resizebox{0.48\textwidth}{!}{%
\begin{tabular}{@{}lcc@{}}
\toprule
\multicolumn{1}{c}{Characteristics to identify gradient obfuscation} & Pass & Fail \\ \midrule
1. One-step attack performs better than iterative attacks & \checkmark &  \\
2. Black-box attacks are better than white-box attacks & \checkmark &  \\
3. Unbounded attacks do not reach 100\% success & \checkmark &  \\
4. Random sampling finds adversarial examples& \checkmark &  \\
5. Increasing distortion bound doesn't increase success & \checkmark &  \\ \bottomrule
\end{tabular}%
}
\end{table}

\paragraph{Inspections of gradient obfuscation.}

The famous gradient obfuscation work \cite{athalye2018obfuscated} enumerates several characteristic behaviors as listed in \cref{tabel:checklist_gradient_obfuscation} which can be observed when the defense method owns gradient obfuscation. Our experiments show that PNI passes each inspection item in \cref{tabel:checklist_gradient_obfuscation}.

\begin{figure}[t]
  \centering
   \includegraphics[width=0.45\textwidth,height=0.3\textwidth]{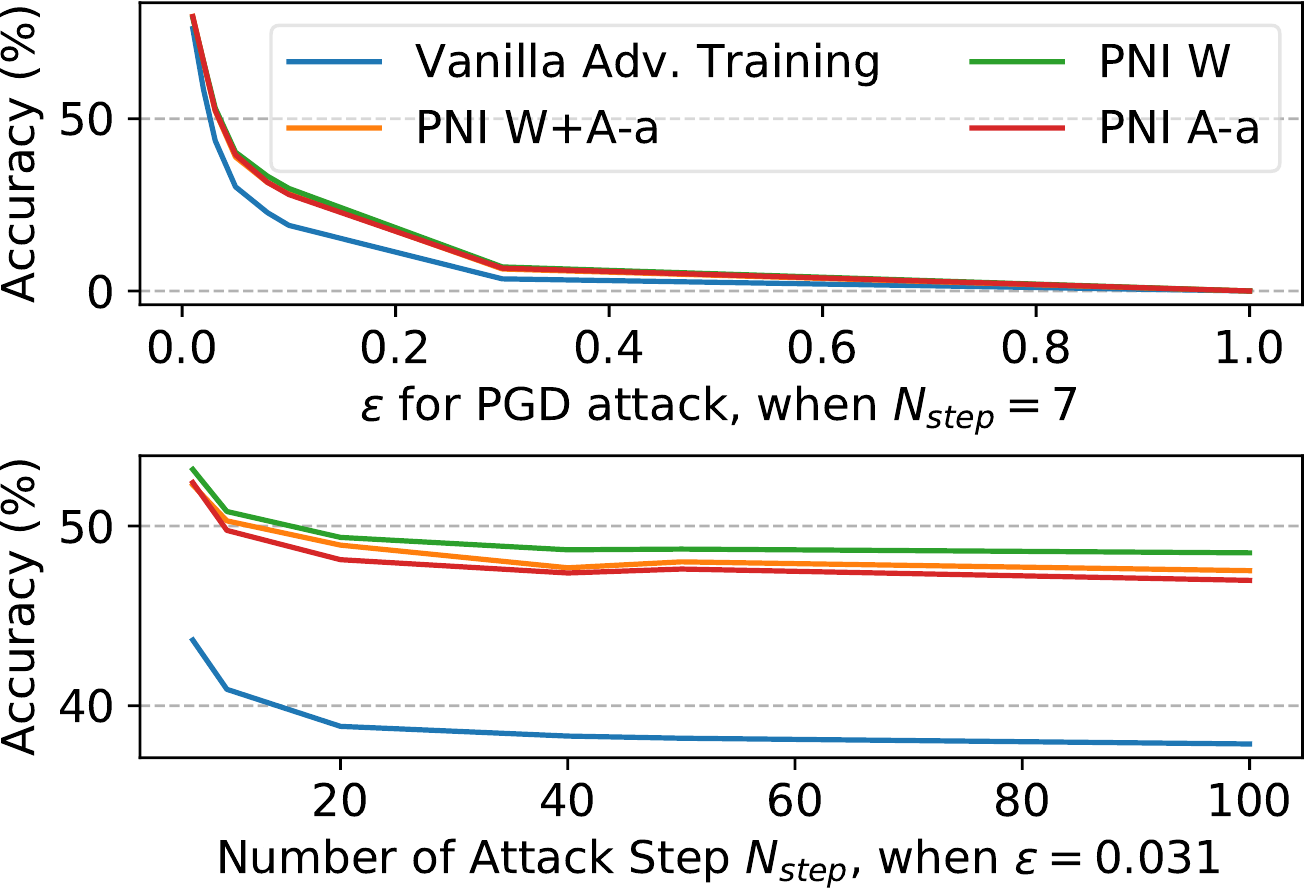}
   \caption{On CIFAR-10 test set, the perturbed-data accuracy of ResNet-18 under PGD attack (Top) versus attack bound $\epsilon$, and (Bottom) versus number of attack steps $N_{step}$}
   \label{fig:accuracy_vs_epsilon_and_steps}
\end{figure}

For item.1, all the experiments in \cref{table:PNI_schemes} and \cref{table:PNI-w_comparison} report that FGSM attack (one-step) performs worse than PGD attack (iterative). For item.2, our black-box attack experiment in \cref{table:PNI_black_box} shows that the black-box attack strength is worse than white-box attack. For items.3, as plotted in \cref{fig:accuracy_vs_epsilon_and_steps}, we run experiments through increasing the distortion bound-$\epsilon$. The result shows that the unbounded attacks do lead to 0\% accuracy under attack. For item.4, the prerequisite is the gradient-based attack (e.g., PGD and FGSM) cannot find the adversarial examples, however the experiments in \cref{fig:accuracy_vs_epsilon_and_steps} reveals that our method still can be broken when increasing the distortion bound. It just increases the resistance against the adversarial attacks, in comparison to the vanilla adversarial training. For item.5, again as shown in \cref{fig:accuracy_vs_epsilon_and_steps}, increasing the distortion bound increase the attack success rate.

\paragraph{PNI does not rely on stochastic gradients.}
As shown in \cref{fig:accuracy_vs_epsilon_and_steps}, gradually increasing the PGD attack steps $N_{step}$ raises the attack strength \cite{madry2018towards}, thus leading to perturbed-data accuracy degradation for both vanilla adversary training and our PNI technique. However, for both cases the perturbed-data accuracy start saturating and do not degrade any further when $N_{step}=40$. If our PNI's success comes from the stochastic gradient which gives incorrect gradient owing to the single sample, increasing the attack steps suppose to eventually break the PNI defense which is not observed here. Our PNI method still outperforms vanilla adversarial training even when $N_{step}$ is increased up to 100. Therefore, we can draw the conclusion that, even if PNI does include gradient obfuscation, the stochastic gradient is not the dominant role in PNI for the robustness improvement.

\section{Conclusion}
In this paper, we present a parametric noise injection technique where the noise intensity can be trained through solving the min-max optimization problem during adversarial training. Through extensive experiments, the proposed PNI method can outperforms the state-of-the-art defense method in terms of both clean-data accuracy and perturbed-data accuracy.

\bibliographystyle{unsrt}
\bibliography{reference}

\end{document}